\documentclass{article}
\usepackage{spconf,amsmath,epsfig}
\usepackage{graphicx}
\usepackage{subeqnarray}
\usepackage{amsmath}
\usepackage{amssymb}
\usepackage{amssymb}
\usepackage{mathtools}
\DeclarePairedDelimiter\norm{\lVert}{\rVert}%
\begin{document}\sloppy

\def\x{{\mathbf x}}
\def\L{{\cal L}}

\title{Convolutional Temporal Attention Model for Video-based Person Re-identification}
%
\name{Tanzila Rahman$^{1*}$ \quad \quad \quad \quad Mrigank Rochan$^{2}$ \quad \quad \quad \quad Yang Wang$^{2}$ \thanks{$^{*}$Work done while at the University of Manitoba.}}
\address{$^{1}$University of British Columbia \quad\quad \quad \quad $^{2}$University of Manitoba\\
 $^{1}$trahman8@cs.ubc.ca,  $^{2}$\{mrochan, ywang\}@cs.umanitoba.ca}

\maketitle

\begin{abstract}
The goal of video-based person re-identification is to match two input videos, so that the distance of the two videos is small if two videos contain the same person. A common approach for person re-identification is to first extract image features for all frames in the video, then aggregate all the features to form a video-level feature. The video-level features of two videos can then be used to calculate the distance of the two videos. In this paper, we propose a temporal attention approach for aggregating frame-level features into a video-level feature vector for re-identification. Our method is motivated by the fact that not all frames in a video are equally informative. We propose a fully convolutional temporal attention model for generating the attention scores. Fully convolutional network (FCN) has been widely used in semantic segmentation for generating 2D output maps. In this paper, we formulate video based person re-identification as a sequence labeling problem like semantic segmentation. We establish a connection between them and modify FCN to generate attention scores to represent the importance of each frame. Extensive experiments on three different benchmark datasets (i.e. iLIDS-VID, PRID-2011 and SDU-VID) show that our proposed method outperforms other state-of-the-art approaches.
\end{abstract}
\section{Introduction}\label{sec:intro}


Person re-identification is an active area of research in computer vision. Earlier work~(e.g. \cite{li2014deepreid, ahmed2015improved}) in this area focuses on image-based re-identification. Given an input query image (called the probe image) of a person, the goal is to find this person from a collection of gallery images. Recently, there has been work~\cite{xu17_iccv,zhou17_cvpr} exploring video-based person re-identification. Compared with images, video-based person re-identification is a more realistic setting in real-world application. In this paper, we focus on video-based person re-identification. Given a query video of a person, we would like to identify the person by matching the query video to a collection of gallery videos.  

\begin{figure}
  \centering
  \includegraphics[height=2in,width=2in]{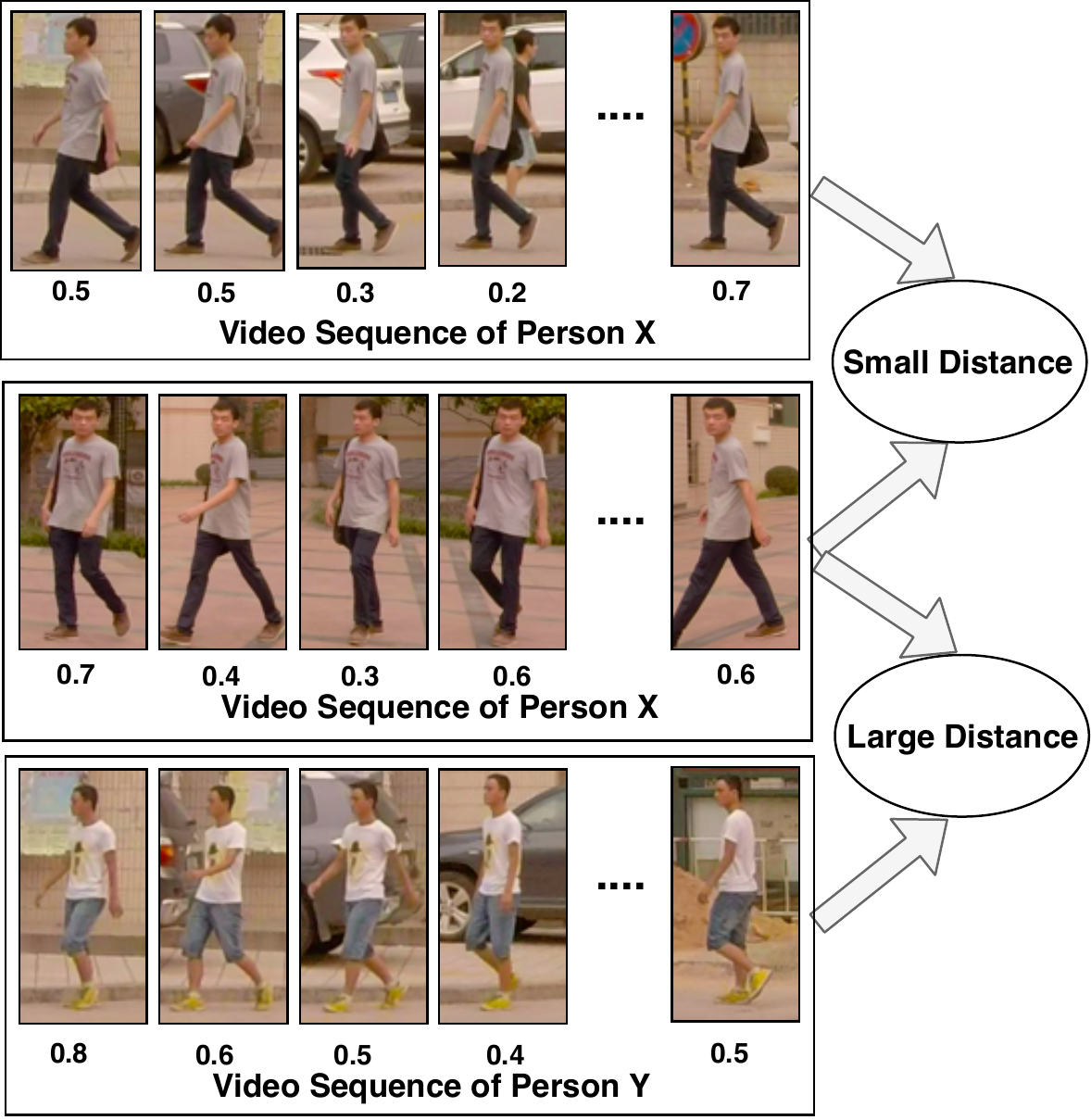}
  \caption{Illustration of video-based person re-identification problem. The problem can be formulated as learning the distance between two input videos. If the two videos contain the same person, their distance should be small. Otherwise, the distance should be large. }
  \label{fig:problem}
\end{figure}

Most work in person re-identification uses some form of metric learning. Given two images (or videos), we would like their distance to be small if they contain the same person, and the distance to be large otherwise. See Figure~\ref{fig:problem} for an illustration in the case of video-based person re-identification. In image-based person re-identification, convolutional neural networks (CNNs) are often used to learn this distance metric. For example, in \cite{li2014deepreid, ahmed2015improved}, CNNs are used to extract features from images in a way that the $L_2$ distance between the extracted features can be used as the distance metric. Most existing approaches \cite{mclaughlin16_cvpr,xu17_iccv,zhou17_cvpr} in video-based person re-identification follow a similar strategy. First, image features are extracted from each frame in the video. These frame-level images features of a video are then aggregated together to form a video-level feature with fixed length. Finally, the distance between two videos are calculated based on their video-level features.

As observed in previous work~\cite{xu17_iccv,zhou17_cvpr}, not all frames in a video are equally informative. For example, the person might be heavily occluded in some frames. Ideally, we would like to pay attention to ``good'' frames when constructing the video-level feature representation. Previous works~\cite{zhou17_cvpr,xu17_iccv} have used recurrent neural networks~(RNNs) to assign a temporal attention score to each frame in a video. The attention score indicates how informative a frame is. Intuitively, the learning algorithm should assign small attention scores to frames where the person is heavily occluded. The video-level feature is obtained by the summation of frame-level features weighted by their attention scores.

Although RNN has been popular in many sequence labeling problems, it has some inherent limitations. The computation involved in RNN is sequential, i.e. we cannot process a frame until all previous frames have been processed. Due to the sequential nature of RNN, it is difficult to take advantage of the GPU hardware and fully parallelize the computation involved in RNN. The same observation has also been made in natural language processing~\cite{BaiTCN2018,gehring2017convolutional}. Some recent works~\cite{BaiTCN2018,gehring2017convolutional} have advocated using convolutional models instead of recurrent models for sequence labeling tasks. Similar to RNN, convolutional models can also capture contextual dependencies in sequential data via their effective receptive field. But different from RNN, convolutional models can better exploit the GPU hardware.

Our contributions include: (1) to the best of our knowledge this is the first attempt to formulate video based person re-identification as semantic segmentation.  We propose a fully convolutional temporal model for generating the attention scores of frames in a video. (2) Unlike previous work~(e.g.~\cite{zhou17_cvpr}) that uses RNN to generate the attentions, our model directly generates attentions based on frame-based features. As a consequence, the computation of the attentions is much simpler and can be easily parallelized. (3) Extensive experiments on three benchmark datasets show that our proposed model outperforms other state-of-the-art methods.

\section{Related Work}\label{sec:related}
The problem of person re-identification can be divided into two categories: image-based and video-based. Previous methods for person re-identification from static images focus on two tasks: (1) extracting features from input images and (2) measuring their similarity or distance metric to determine whether the images belong to same person or not. Discriminative features play a vital role to handle environmental and view points changes.~\cite{wang2014unsupervised} and~\cite{zhao2013unsupervised} propose methods that consider patch appearance statistics to localize important part of an individual person. In \cite{gray2008viewpoint}, an ensemble of spatial and color information is used to increase viewpoint variations. After extracting features from images, distance metric learning is applied to increase the distance between different persons. For the same persons distance should decrease. In deep learning, both feature extraction and distance metric learning are applied in an end-to-end fashion for re-identifying person.

In recent years, researchers start to pay more attention to video-based person re-identification, partly because this is a more realistic setting in real-world applications. Previous method~\cite{simonnet2012re} consider frame level similarity to identify the same person. Recently, deep learning based approaches are gaining popularity for video re-identification. Most of them use the siamese architecture where each branch contains RNN to capture temporal information. McLaughlin et al.~\cite{mclaughlin16_cvpr} propose a method which collects temporal information using optical flows, recurrent neural network (RNN) and temporal pooling layers. Following~\cite{mclaughlin16_cvpr}, Xu et al.~\cite{xu17_iccv} propose a Spatial and Temporal Attention Pooling Network (ASTPN) for learning interdependence information.  Our work is motivated by the recent success of attention-based models~\cite{bahdanau15_iclr, yin16_tacl}. In this paper, we generate an attention score for each frame which indicates the importance of that frame within the input video sequence. The main contribution of our method is establishing the connection between video based person re-identification and semantic segmentation. Instead of RNN, here we adopt fully convolution network (FCN) for generating attention over the video frames. 

\section{Our Approach}\label{sec:approach}
\begin{figure*}
\begin{center}
  \includegraphics[width=14.8cm, height=3.9cm]{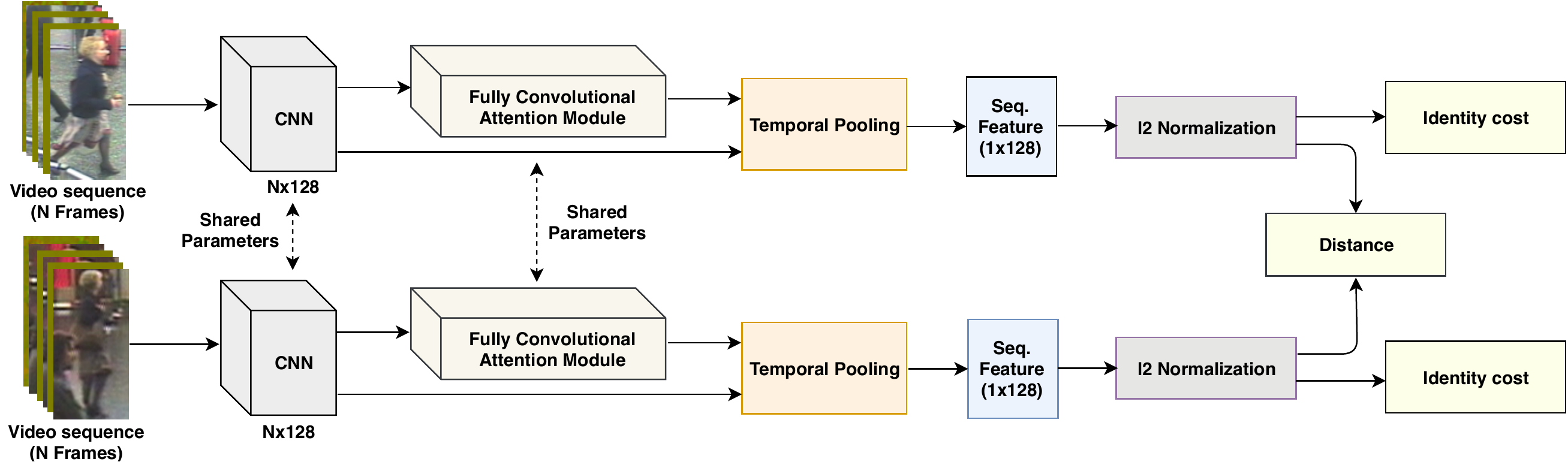}
\end{center}
   \caption{Illustration of our proposed network architecture for video-based person re-identification. The network takes a pair of input video sequences where each sequence consists of $N$ number of frames. Each frame in a video is passed to a Convolutional Neural Network (CNN) to generate a 128-dimensional frame level feature. In other words, each video is represented as a feature matrix of dimensions $N \times 128$. We then use fully convolutional attention module that takes the $N\times 128$ feature matrix and generates a $N\times 1$ vector of attention scores for frames in the video. Then we use the attention scores to re-weight frame-level features and produce a 128-dimensional video-level feature vector using a temporal pooling layer. The video-level feature vector is then normalized and used to compute the distance of two input videos. We use a squared hinge loss on the distance for learning. During training, we also use the video-level feature vector to classify the identity of the person in a video and use a standard softmax loss for the classification.}
\label{fig:block_diagram}
\end{figure*}

Our proposed approach uses a Siamese network architecture~(see Figure~\ref{fig:block_diagram}). Our network takes a pair of input video sequences as its input. It outputs a scalar value indicating how like these two videos contain the same person. The network architecture has two identical branches with shared parameters. Each branch of the network takes a video sequence as the input and extract per-frame features using CNN. Then we compute attention score for each frame using Fully Convolution Network (FCN). The attention score indicates the importance of the frame for the re-identification task. The video-level feature representation is obtained by aggregating the frame-level feature weighted by the corresponding attention score on the frame. Finally, the video-level features of the two input videos are used to compute their distance for re-identification.

\subsection{Frame Feature Extraction Module}\label{sec:cnn}
Following \cite{mclaughlin16_cvpr}, we use both RGB color and optical flow channels for extracting frame-level features. Color channels give information about a person's appearance while optical flows give motion related information. As a preprocessing step, first we convert input video frames from RGB to YUV color space. Each color channel is then normalized to have a zero mean and unit variance. Both vertical and horizontal optical flow channels (e.g. $\Gamma_x$ and $\Gamma_y$ respectively) for each frame is calculated using Lucas-Kanade algorithm \cite{lucas1981_iterative}.  In the end, each input frame is represented as a tensor of dimensions $56 \times 40 \times 5$ where the 5 channels correspond to 3 color channels and 2 optical flow channels. We use a CNN architecture similar to \cite{mclaughlin16_cvpr} to extract frame-level features. The CNN architecture consists of three stages of convolution, max-pooling, and nonlinear ($\mathrm{tanh}$) activations. Each convolution filter uses $5\times 5$ kernels with $1\times 1$ stride and $4\times 4$ zero padding. In the end, CNN produces a 128-dimensional feature vector (i.e. $\mathbf{z}_i\in\mathbb{R}^{128}$) to represent each frame $\mathbf{z}_i$ ($i=1,2,...,N$) in the input video.
 
\subsection{Fully Convolutional Attention Module}\label{sec:FCN_attention}
 Motivated by the recent attention-based models (e.g. \cite{bahdanau15_iclr, yin16_tacl}), we introduce an attention-based approach for re-identifying person from video sequences. The attention-based approach is inspired by the visual processing of human brains which often pay attention on discriminative regions of different frames instead of whole video when try to re-identify persons \cite{xu17_iccv}. In this paper, we focus on temporal attentions by formulating it as a sequence labeling problem like semantic segmentation. In particular, we adopt a 1D temporal version of the Fully Convolution Network (FCN)~\cite{long2015fully} to generate the temporal attentions. FCN is a widely used network architecture for semantic segmentation. Let $X\in\mathbb{R}^{H\times W\times 3}$ be an input image with $H\times W$ spatial dimensions and 3 color channels. FCN first uses an encoder network to extract a feature map from the image $X$. The feature map is then used to produce an output map $Y$ with the same spatial dimension of the input image, i.e. $Y\in\mathbb{R}^{H\times W}$. Each entry of the output map $Y$ represents the semantic label at the corresponding pixel location in the image. In summary, FCN processes a 2D image (with 3 channels) and produces a 2D map as the output.

\begin{figure}[t!]
\begin{center}
  \includegraphics[width=5cm, height=10cm]{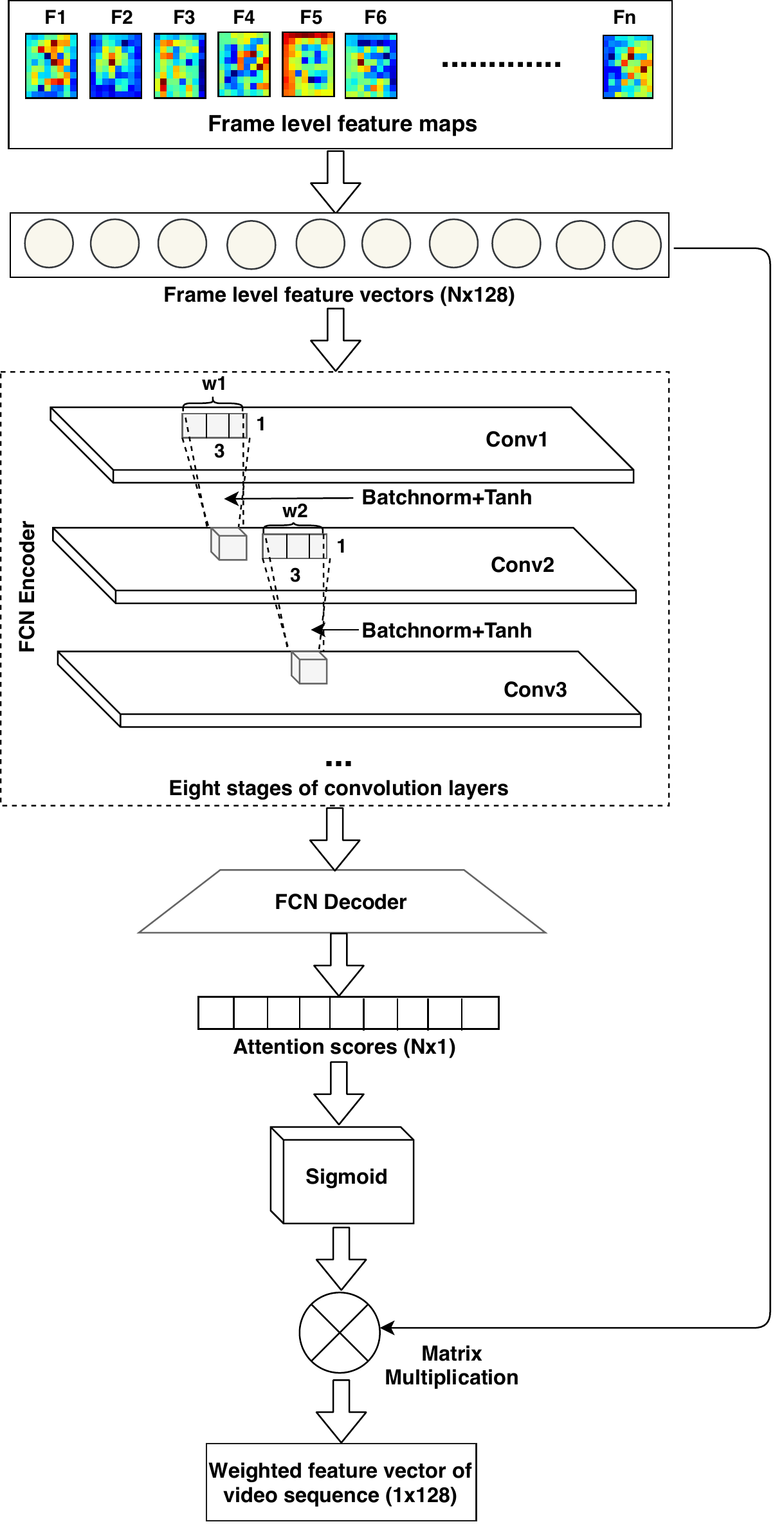}
\end{center}
   \caption{Detailed architecture of our fully convolutional attention module for generating temporal attentions on frames in a video. The input of the attention module is the feature vectors generated by the frame-level CNN (see Sec.~\ref{sec:cnn}) where each frame is represented as a 128-dimensional vector. We adopt fully convolution network (FCN) to take an input as the frame features ($N \times 128$) and generate $N\times 1$ attention scores. Here, $w_1$ and $w_2$ are used to represent window size of the convolutions. The attention scores are then normalized using the Sigmoid activation function and multiplied with the frame-level feature vectors to generate a weighted feature for the entire video. }
\label{fig:FCN_architecture}
\end{figure} 

The main insight of this paper is as follows. Suppose each frame in a video has been processed (see Sec.~\ref{sec:cnn}) and represented as a 1-dimension feature vector. We can treat a video with $N$ as a 1D input ``image'' with 128 channels. If we want to assign an attention score to each frame, we can treat these attention scores as a $N\times 1$ output map. In other words, we can adopt FCN to generate attention scores for frames in a video by making the following modifications: 1) instead of taking an input image of size $H\times W$ with 3 channels, we now take $N\times 1$ input with 128 channels; 2) instead of producing $H\times W$ output map, we now produce $N\times 1$ output map; 3) instead of 2D convolution and 2D pooling, we will perform 1D convolution and 1D pooling operations. We call it fully convolutional attention module.

Figure~\ref{fig:FCN_architecture} shows the detailed architecture of the fully convolutional temporal attention module. Each frame of an input video is represented as a 128-dimensional frame-level feature vector (see Sec.~\ref{sec:cnn}). The sequence of frame-level feature vectors are passed to the fully convolutional attention module to generate temporal attention scores. In this paper, we adopt FCN32s \cite{long2015fully} which is the basic fully convolutional network used in semantic segmentation. Input to the attention module is a tensor of dimension $N\times 128$ where $N$ corresponding to the number of input video frames. The output is a $N\times 1$ vector. The output is followed by a Sigmoid function. In the end, we obtain an attention score $\lambda_i$ for each feature vector $\mathbf{z}_i$ where $i$ represents $i$-th frame in video sequence. We can express the attention scores using following equation:

\begin{subeqnarray}
&& \alpha_1,\alpha_2,...,\alpha_{N}=FCN(\mathbf{z}_1, \mathbf{z}_2,...,\mathbf{z}_N)\\
&& {\lambda_i}=\frac{1}{1+\exp(-{\alpha_i})}, \ \ \textrm{where } i=1,2,...,N\label{eq:attention}
\end{subeqnarray}
where $FCN(\cdot)$ represent the FCN network that takes frame-level features $\mathbf{z}_1, \mathbf{z}_2,...,\mathbf{z}_N$ and produce a vector $[\alpha_1,\alpha_2,...,\alpha_{N}]$ of unnormalized attention scores. This is followed by a Sigmoid function to produce normalized attention scores $\lambda_i$ ($i=1,2,...,N$).

The frame-level attention scores $\lambda_i$ ($i=1,2,...,N$) are then combined with the per-frame feature vectors $z_i$ ($i=1,2,...,N$) via an attention pooling to generate a weighted feature vector $\mathbf{\gamma}$ as follows:
\begin{equation}
 \mathbf{\gamma}=\sum_{i=1}^{N}{\lambda_i}{\mathbf{z}_i}\label{eq:video_feature}
\end{equation}
Here $\mathbf{\gamma}$ can be treated as a feature vector of entire video sequence. We have also tried Softmax in Eq.~\ref{eq:attention} and found that it does not perform as well as Sigmoid. This is consistent with the observation in previous work~\cite{zhao2017_deeply}, 

In our experiments, we find that the video-level feature obtained by a regular temporal pooling layer~\cite{mclaughlin16_cvpr} (i.e. $\frac{1}{N}\sum_{i=1}^{N}\mathbf{z}_i$) can also help to improve the performance. So our final video-level feature $F$ is the average of the feature vectors obtained from the attention pooling and the regular temporal pooling. We apply $l_2$ normalization on $F$ in the end.



\subsection{Model Learning}
In this section, we describe the details of learning the parameters of our model. Let ${F}_1$ and ${F}_2$ be the feature vectors of two input videos from the Siamese network. Following \cite{mclaughlin16_cvpr,xu17_iccv}, we calculate Euclidean distance between the feature vectors and apply the squared hinge loss($Loss_{hinge}$) as follows:
\begin{equation}
  \label{eq:hinge}
 {\xi_{hinge}=\begin{cases}
      {\frac{1}{2}\norm{{{F}_1}-{{F}_2}}^{2} }, & {X_1}={X_2} \\
      \frac{1}{2}[\max(0,m-{\norm{{{F}_1}-{{F}_2}} })]^2, & {X_1}\neq{X_2}
    \end{cases}}
\end{equation}
where the hyper-parameter $m$ represents the margin of separating two classes in $\xi_{hinge}$. Here we use $X_1$ and $X_2$ to represent the identities of the persons from two input videos. The idea is that if the two videos contain the same person (i.e. ${X_1}={X_2}$), the distance between the feature vectors should be small. Otherwise, the distance should be large if the persons are different (i.e. ${X_1}\neq{X_2}$).

Similar to \cite{mclaughlin16_cvpr}, we also use another loss (i.e. identity loss $Loss_{id}$) to each branch of the Siamese network to predict the person's identity. We use a linear classifier to predict one of the person's identity from the feature vector extracted through each branch of the Siamese network. We then apply a Softmax loss over the prediction for each Siamese branch. The final loss is the combination of two identity losses (i.e. $\xi_{id1}$ and $\xi_{id2}$) from each Siamese branch and the hinge loss as follows:
\begin{equation}
\label{eq:loss_final}
{\xi_{final}=\xi_{id1}+\xi_{hinge}+\xi_{id2}}
\end{equation}

We use stochastic gradient decent to optimize the loss function define in Eq.~\ref{eq:loss_final}. After training, we remove all loss functions including the identity and hinge losses. During testing, we only use the feature vectors to compute the distance between two input videos for re-identification.

\section{Experiments}\label{sec:experiment}
In this paper, we use three benchmark datasets (i.e. iLIDS-VID \cite{wang14_eccv}, PRID-2011 \cite{hirzer11_scia} and SDU-VID \cite{liu2015spatio}) for evaluating our proposed method.
We first describe the experiment setup and some implementation details~(Sec.~\ref{sec:implementation}). Then, we present the experimental results and compare with previous work~(Sec.~\ref{sec:results}).

\begin{figure}
  \begin{center}
    \setlength\tabcolsep{1pt}
    \begin{tabular}{ccccccccccc}
   
      \includegraphics[height=0.5in]{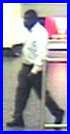}&
      \includegraphics[height=0.5in]{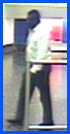}&
      \includegraphics[height=0.5in]{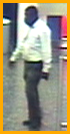}&
      \includegraphics[height=0.5in]{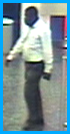}&
      \includegraphics[height=0.5in]{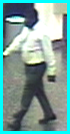}&
      \includegraphics[height=0.5in]{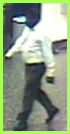}&
      \includegraphics[height=0.5in]{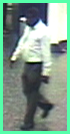}&
      \includegraphics[height=0.5in]{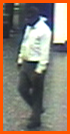}&
      \includegraphics[height=0.5in]{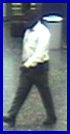}&
      \includegraphics[height=0.5in]{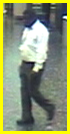}&
      \includegraphics[height=0.5in]{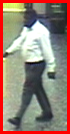}\vspace{1pt}\\ 
    
      \includegraphics[height=0.5in]{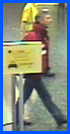}&
      \includegraphics[height=0.5in]{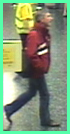}&
      \includegraphics[height=0.5in]{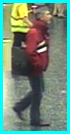}&
      \includegraphics[height=0.5in]{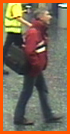}&
      \includegraphics[height=0.5in]{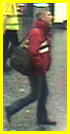}&
      \includegraphics[height=0.5in]{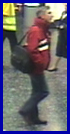}&
      \includegraphics[height=0.5in]{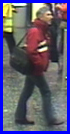}&
      \includegraphics[height=0.5in]{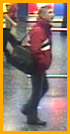}&
      \includegraphics[height=0.5in]{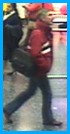}&
      \includegraphics[height=0.5in]{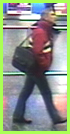}&
      \includegraphics[height=0.5in]{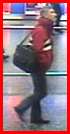}\vspace{1pt}\\
      
      \multicolumn{11}{c}{\includegraphics[height=0.2in]{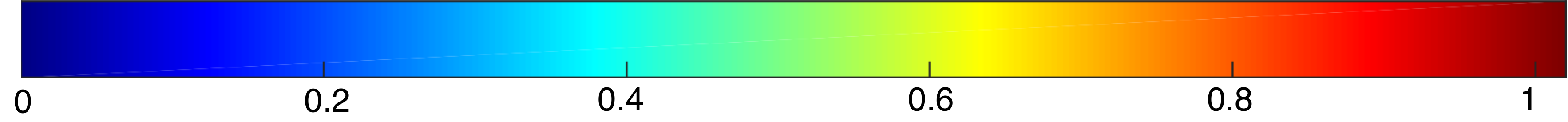}}\vspace{1pt}\\
    \end{tabular}
\end{center}
   \caption{Visualization of learned attention scores of our approach. Each row shows selected frames in a video. The color of the box enclosing the frame indicates the corresponding attention score of the frame. Warm colors correspond to high attention scores. We can see that frames with high attention scores tend to have less background clutters and occlusions.}
\label{fig:attention_score}
\end{figure}

\subsection{Setup and Implementation Details}\label{sec:implementation}
We follow the experimental protocol of McLaughlin et al. \cite{mclaughlin16_cvpr} on the iLIDS-VID and PRID-2011 datasets. We randomly split the datasets into two equal parts: one part is used for training and the remaining part is used for testing. We repeat all experiments 10 times for stable result. We use the Cumulative Matching Characteristics (CMC) curve. We use equal numbers of positive and negative samples during training to alleviate the effect of class imbalance. We use $m=2$ as the value of hyper-parameter in the hinge loss. The network is trained for 1400 epochs. We keep the batch size as one. The initial learning rate is set as $1e^{-4}$. For the iLIDS-VID dataset, we decrease the learning rate by a factor of 10 after 1300 epochs. On the PRID-2011 dataset, we decrease the learning rate after 800 and 1100 epochs by a factor of 10. For the SDU-VID dataset, we follow the experimental protocol of Zhang et al.\cite{zhang2017learning}. For this dataset, we decrease the learning rate by a factor of 10 after 1200 epochs.

\subsection{Experimental Results}\label{sec:results}
We show the experimental results and comparisons with previous methods on the three datasets in Table~\ref{table:iLIDS_VID}, Table~\ref{table:PRID-2011} and Table~\ref{table:SDU-VID}, respectively. We can see that our proposed methods significantly outperforms previous methods in terms of the rank-1 CMC accuracy on all datasets. The comparison with \cite{zhou17_cvpr} is particularly interesting since \cite{zhou17_cvpr} uses a similar temporal attention approach. The difference is that \cite{zhou17_cvpr} uses RNN to generate attention scores, while our method uses a fully convolutional network (FCN) to generate attention scores. This shows that convolutional models provide a competitive alternative to RNN for temporal attentions. In addition to temporal attentions, \cite{zhou17_cvpr} additionally uses a recurrent model to generate spatial attentions. In contrast, our model only uses temporal attentions and is much simpler, yet still achieves much better performance. Figure~\ref{fig:attention_score} shows some visualizations of the attention scores on some frames within same videos. Intuitively, frames with high attention scores tend to be the ones with less background clutters and occlusions.

\begin{table}[h!]
\footnotesize
   \begin{center}
  \begin{tabular}{|l|c|c|c|c|}
  \hline
     Method & Rank-1 & Rank-5 & Rank-10 & Rank-20 \\
    \hline
    Ours & {\bf 64} & \textbf{92} & \textbf{96} & {\bf 98}\\
    \hline
    Xu et al.\cite{xu17_iccv} & 62 & 86 & 94 & 98\\
    \hline
    Zhou et al.\cite{zhou17_cvpr} & 55.2 & 86.5 & - & 97.0\\
    \hline
    McLaughlin et al.\cite{mclaughlin16_cvpr} & 58 & 84 & 91 & 96 \\
    \hline
    Yan et al.\cite{yan16_eccv}& 49 & 77 & 85 & 92\\
    \hline
  \end{tabular}
  \end{center}
  \caption{Performance comparison of our proposed method with other state-of-the-art on the iLIDS-VID dataset in terms of CMC(\%) ranking metric.}
  \label{table:iLIDS_VID}
\end{table}

\begin{table}[h!]
\footnotesize
   \begin{center}
  \begin{tabular}{|l|c|c|c|c|}
  \hline
     Method & Rank-1 & Rank-5 & Rank-10 & Rank-20 \\
    \hline
    Ours & \textbf{90} & \textbf{98} & 98 & 99\\
    \hline
    Xu et al.\cite{xu17_iccv} & 77 & 95 & \textbf{99} & 99\\
    \hline
    Zhou et al.\cite{zhou17_cvpr} & 79.4 & 94.4 & - & \textbf{99.3}\\
     \hline
    McLaughlin et al.\cite{mclaughlin16_cvpr} & 70 & 90 & 95 & 97 \\
    \hline
    Yan et al.\cite{yan16_eccv}&64 & 86 & 93 & 98\\
    \hline
  \end{tabular}
  \end{center}
  \caption{Performance comparison of our proposed method with other state-of-the-art on the PRID-2011 dataset in terms of CMC(\%) ranking metric.}
  \label{table:PRID-2011}
\end{table}

\begin{table}[h!]
\footnotesize
   \begin{center}
  \begin{tabular}{|l|c|c|c|c|}
  \hline
     Method & Rank-1 & Rank-5 & Rank-10 & Rank-20 \\
    \hline
    Ours & \textbf{87} & \textbf{97} & 98 & \textbf{100}\\
    \hline
    Zhang et al.~\cite{zhang2017learning} & 85.6 & 97 & \textbf{98.3} & 99.6\\ 
    \hline
    RNN~\cite{mclaughlin16_cvpr} & 75.0 & 86.7 & - & 90.8\\
    \hline
    STA+KISSME~\cite{liu2015spatio} & 73.3 & 92.7 & 95.3 & 96.0\\
    \hline
    Liu et al.~\cite{liu2015spatio} & 62.0 & 81.3 & - & 92.7\\
    \hline
  \end{tabular}
  \end{center}
  \caption{Performance comparison of our proposed method with other state-of-the-art on the SDU-VID dataset in terms of CMC(\%) ranking metric.}
  \label{table:SDU-VID}
\end{table}

\subsection{Cross-Dataset Testing}\label{sec:cross_data}
In this section, we perform cross-dataset testing to further test the generalizability of our method. A system usually performs better when it is trained and tested on the same dataset due to the data bias. But in real-world applications, test data are usually totally different from the training data used during learning. To estimate the real-world performance of a system, a better way is to do cross-dataset testing where we train the model using one dataset and test the model using a completely different dataset. Following previous work~\cite{mclaughlin16_cvpr}, we use 50\% of the larger and more challenging iLIDS-VID dataset for training our network, and use 50\% of the PRID-2011 dataset for testing. For this type of cross-dataset testing, previous methods \cite{mclaughlin16_cvpr, xu17_iccv} have used two different settings: single-shot re-identification and multi-shot re-identification. In the single-shot re-identification, only one frame of a video is used. But in our case, we can not apply the single-shot setting as our method generates attention scores based on frame features and later combines frame-level feature vectors to produce the video-level features. If we only use one image, the method is equivalent to simply rescaling a frame-based feature vector by a constant and then use it for re-identification. We compare the results with other multi-shot cross-dataset testing scenario in Table~\ref{table:multi_cross_dataset}. Again, our method outperforms other methods in terms of CMC ranking accuracy. 

\begin{table}[h!]
\footnotesize
   \begin{center}
  \begin{tabular}{|l|c|c|c|c|c|}
  \hline
     Method & Trained on & Rank-1 & Rank-5 & Rank-10 & Rank-20 \\
    \hline
    Ours & iLIDS-VID & {\bf 32} & {\bf 60} & {\bf 72} & {\bf 86}\\
    \hline
    \cite{xu17_iccv} & iLIDS-VID & 30 & 58 & 71 & 85\\
    \hline
    \cite{mclaughlin16_cvpr} & iLIDS-VID & 28 & 57 & 69 & 81 \\
    \hline
  \end{tabular}
  \end{center}
  \caption{CMC Rank accuracy (\%) using cross dataset testing (using multi-shot re-identification) on the PRID-2011 dataset. The model is trained on the iLIDS-VID dataset.}
  \label{table:multi_cross_dataset}
\end{table}

\section{Conclusion}\label{sec:conclude}
In this paper, we have proposed a temporal attention approach for video-based person re-identification. The novelty of our model is that we use a fully convolutional model for generating the temporal attentions. Fully convolutional network (FCN) has been widely used to produce 2D output (e.g. in semantic segmentation). Our proposed method modifies traditional FCN to produce a 1D output (i.e. temporal attentions). Through extensive experiments, we have demonstrated that the proposed method outperforms existing state-of-the-art video-based person re-identification methods.


\hfill \break
\textbf{Acknowledgments: }The authors acknowledge financial support from NSERC, MGS and UMGF funding. We also thank NVIDIA for donating some of the GPUs used in this work.

\bibliographystyle{IEEEbib}
\bibliography{egbib}

\end{document}